\documentclass[letterpaper,journal]{IEEEtran}
\usepackage{amsmath,amsfonts,amssymb}
\usepackage{array}
\usepackage[T1]{fontenc}
\usepackage{textcomp}
\usepackage{stfloats}
\usepackage{url}
\usepackage{verbatim}
\usepackage{graphicx}
\usepackage{gensymb,soul}
\usepackage{booktabs}
\usepackage[caption=false,font=footnotesize]{subfig}
\usepackage{epsfig} 
\usepackage{cite}
\usepackage{bm}
\usepackage[colorlinks,linkcolor=blue,citecolor=blue]{hyperref}

\usepackage{xcolor}
\usepackage{ifthen}
\usepackage{makecell}
\usepackage{algpseudocode}
\usepackage{threeparttable}
\usepackage[linesnumbered,ruled]{algorithm2e}

\DeclareMathAlphabet{\mathcal}{OMS}{cmsy}{m}{n}
\DeclareSymbolFont{largesymbols}{OMX}{cmex}{m}{n}

\usepackage{multirow}

\begin{document}

\title{SAMP: Spatial Anchor-based Motion Policy for Collision-Aware Robotic Manipulators}

\author{Kai Chen, Zhihai Bi, Guoyang Zhao, Chunxin Zheng, Yulin Li, Hang Zhao, and Jun Ma, \textit{Senior Member, IEEE}
\thanks{Kai Chen, Zhiahai Bi, Guoyang Zhao, Chunxin Zheng, and Hang Zhao are with the Robotics and Autonomous Systems Thrust, The Hong Kong University of Science and Technology (Guangzhou), Guangzhou, China (e-mail: kchen916@connect.hkust-gz.edu.cn). }
\thanks{Yulin Li and Jun Ma are with the Robotics and Autonomous Systems Thrust, The Hong Kong University of Science and Technology (Guangzhou), Guangzhou, China, and the Division of Emerging Interdisciplinary Areas, The Hong Kong University of Science and Technology, Hong Kong SAR, China (e-mail: jun.ma@ust.hk).}
}

\maketitle

\begin{abstract}
Neural-based motion planning methods have achieved remarkable progress for robotic manipulators, yet a fundamental challenge lies in simultaneously accounting for both the robot’s physical shape and the surrounding environment when generating safe and feasible motions. Moreover, existing approaches often rely on simplified robot models or focus primarily on obstacle representation, which can lead to incomplete collision detection and degraded performance in cluttered scenes.
To address these limitations, we propose spatial anchor-based motion policy (SAMP), a unified framework that simultaneously encodes the environment and the manipulator using signed distance field (SDF) anchored on a shared spatial grid. SAMP incorporates a dedicated robot SDF network that captures the manipulator’s precise geometry, enabling collision-aware reasoning beyond coarse link approximations. These representations are fused on spatial anchors and used to train a neural motion policy that generates smooth, collision-free trajectories in the proposed efficient feature alignment strategy. Experiments conducted in both simulated and real-world environments consistently show that SAMP outperforms existing methods, delivering an 11\,\% increase in success rate and a 7\,\% reduction in collision rate. These results highlight the benefits of jointly modelling robot and environment geometry, demonstrating its practical value in challenging real-world environments.
The source code will be released on website\footnote{Available at:~\href{https://github.com/kkkkkaiai/SAMP.git}{github.com/kkkkkaiai/SAMP}}. 
\end{abstract}


\section{Introduction}



Efficient computation of collision-free motions remains a fundamental challenge in robotic planning. Traditional motion planning methods for robots typically operate in configuration space, which execute a projection of obstacles in the workspace and perform collision checks during path search with sampling-based methods or trajectory optimization \cite{path_check1}, \cite{cuRobo}, \cite{path_check2}. Although rigorous, the projecting process is computationally expensive in high-dimensional or complex environments, which significantly slows down the planning process.
To alleviate this computational burden, some approaches use simplified geometric approximations in the workspace \cite{reactive_ds}. Such a compromise routine often results in overly conservative behavior or even infeasible paths in cluttered scenarios.

To overcome these limitations, recent research has shifted toward neural-based methods that aim to imitate expert strategies by implicitly encoding environmental constraints, thus offering a promising pathway toward more efficient and generalizable motion planning in complex environments \cite{motionpolicynetwork}, \cite{constrain_mani}, \cite{constrain_non}, \cite{graph_motion}, \cite{constrain_nmp}. These approaches formulate motion planning as a learning problem, where neural networks are trained to map the robot’s current and goal states, together with a representation of the environment, to feasible actions or trajectory segments without requiring explicit collision checks at each step. Although these methods have demonstrated improved planning efficiency, their performance is highly dependent on the choice of environment representation. Critically, many approaches still rely on specific priors or assumptions about obstacle characteristics, which are similar to those adopted in traditional methods, limiting generalization to novel or previously unseen environments \cite{nn_sdf}, \cite{simp_geo}.

Recent neural-based motion planning methods have explored various strategies for encoding environmental information. Some methods rely on parameterizing obstacles with regular, predefined geometric models \cite{zucker2013chomp}, \cite{simpnet}. While effective in structured settings, these representations struggle to generalize to complex or unstructured environments. More flexible alternatives directly utilize raw point clouds acquired from sensors \cite{MPN}, circumventing explicit geometric assumptions and improving adaptability. Further advances incorporate semantic labels into point cloud representations \cite{elevation_semantic}, \cite{neuralMp} for more precise obstacle representation, though often at the cost of increased training complexity. Another line of work represents the workspace as a binary occupancy map \cite{ntfields}, indicating obstacle locations at a fixed resolution. Although this approach is broadly applicable and straightforward, grid-based representations inevitably suffer from discretization errors and resolution limitations, which undermine accuracy in intricate scenarios.

To address the above issues, we propose SAMP, a spatial anchor-based motion planning framework. SAMP constructs a grid of anchor points across the robot's workspace and leverages neural implicit representations to infer signed distance field (SDF) values at each location, yielding a concise yet expressive description of environmental geometry. To capture robot geometry, SAMP samples points at varying distances around robot links and evaluates them using a dedicated robot SDF network, providing precise geometric features for collision-aware reasoning. At the framework level, the main innovation of SAMP is an effective feature alignment strategy, which combines the environmental and robot SDF features at shared spatial anchor points, and employs a training pipeline based on this comprehensive representation to enhance planning accuracy and generalization. In summary, our main contributions are as follows:

\begin{enumerate}
\item[1)] A unified neural motion planning framework, termed as SAMP, is proposed to leverage spatial anchor points with SDF representations. This framework achieves dense and continuous modeling of environmental geometry, which significantly reduces the learning complexity in cluttered environments.

\item[2)] A novel sampling strategy is further presented to incorporate robot geometry by evaluating points at varying distances around robot links via a dedicated robot SDF network, and this provides precise modeling of robot-specific constraints for collision-aware motion planning.

\item[3)] The proposed training pipeline fused with an effective feature fusion strategy that seamlessly integrates environmental and robot SDF representations, allowing the neural motion policy to jointly optimize for workspace understanding and robot geometry.

\item[4)] Extensive experiments demonstrate that SAMP substantially outperforms established neural motion planning methods in both efficiency and success rate, particularly in high-complexity and previously unseen scenarios.
\end{enumerate}

\section{Related Works}

\subsection{Geometry Representation for Collision-Free Motion Generation}
Accurate environment modeling fundamentally governs motion planning feasibility. Point cloud data suffers from inherent limitations that hinder its effectiveness \cite{motionpolicynetwork}, \cite{neuralMp}, \cite{rmmi}, particularly the substantial structural variance that leads to training instability in neural network integration. While occupancy voxel representations address structural consistency issues, they introduce discretization errors that degrade geometric precision \cite{aggressive_voxel}. SDF provides a more effective alternative through continuous formulation, although traditional voxel-based SDF implementations remain constrained by the memory demands of high-resolution grids \cite{voxblox}, \cite{voxfield}, \cite{nvblox}. Recent neural implicit representations overcome these limitations by learning continuous SDF approximations that enable both efficient collision queries and improved accuracy, as demonstrated by the reconstruction performance of iSDF and subsequent refined frameworks \cite{iSDF2022}, \cite{shine}, \cite{g2sdf}. Building upon these advances, our work introduces spatial anchor-based environmental representations for neural planning, enabling more effective utilization of detailed obstacle descriptions through structured encoding.

The representation of robot geometry constitutes an equally critical factor as environmental modeling in motion planning systems, jointly determining the quality of motion planning results. While current systems predominantly rely on simplified geometric primitives such as spheres, capsules, or bounding boxes to prioritize computational efficiency \cite{simple_robot_geo}, these approximations inevitably compromise modeling fidelity. Alternative approaches using dense point clouds can represent robot geometry more accurately, but introduce new challenges in computational tractability and robustness \cite{fishman2024avoid}. Recent neural implicit representations have demonstrated promising approximation of robot geometry, but they continue to face a fundamental challenge in proper usage for neural motion planning \cite{compositeSDF}, \cite{percep_local}. To bridge this gap, we propose a learnable SDF framework that better captures robot geometry representation and a joint grid sampling method to align the robot and environmental SDF for encoding features.

\subsection{Neural Motion Planning Framework}

The limitations of classical motion planning approaches have motivated the development of neural-based frameworks that aim to directly map sensory observations and robot states to feasible trajectories. A common paradigm is to leverage imitation learning from expert demonstrations, allowing neural networks to predict collision-free paths while bypassing the expensive sampling or optimization procedures required by conventional planners \cite{constrain_nmp}, \cite{MPN}, \cite{constrained_neuralmp}. Beyond simple regression of trajectories, recent works have explored a variety of architectures to improve planning performance and generalization. For example, graph neural networks have been adopted to encode robot kinematics and environmental topology, enabling reasoning over high-dimensional configuration spaces~\cite{graph_motion}. Diffusion-based generative models and conditional variational autoencoders have been proposed to sample diverse, multimodal solutions and avoid local minima~\cite{simpnet,mp_diffusion}. In addition, several approaches transition from direct state supervision to learning adaptive trajectory distributions along expert paths, which better capture motion smoothness and long-horizon feasibility~\cite{ATOA}. In this work, we aim to integrate environment and robot information via spatial anchors and SDF, improving planning accuracy and generalization.

\begin{figure*}
    \centering
    \includegraphics[width=0.9\linewidth]{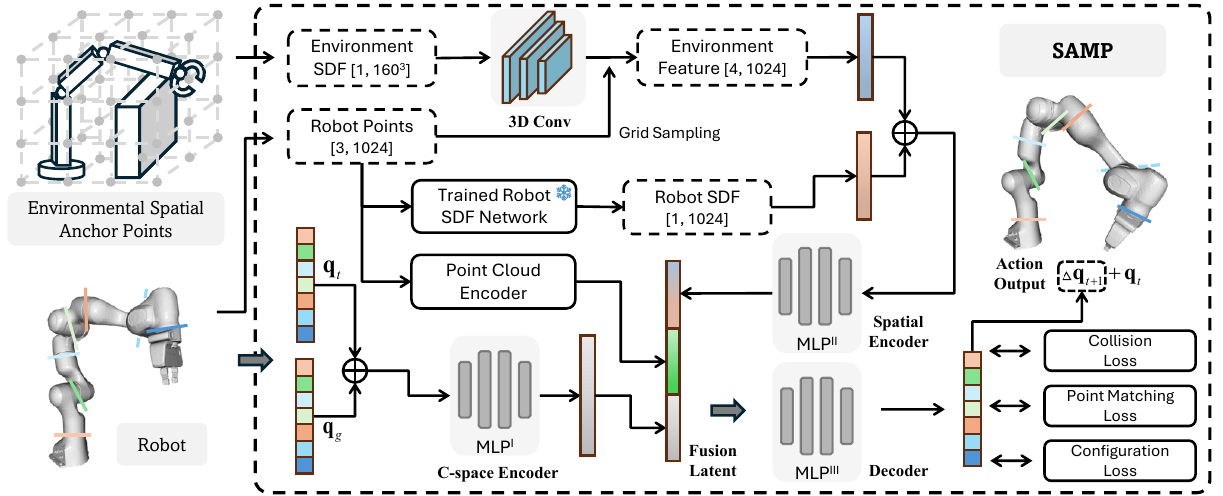}
    \caption{
Overview of the proposed SAMP framework. The system takes as input the robot’s current and target joint configurations, along with environment features. The features are first extracted by processing the environment SDF with a 3D convolutional network, and then grid sampling is performed to obtain features at the robot point locations. In parallel, the pre-trained robot SDF network and a point cloud encoder jointly encode the robot’s geometry and assess collision risks. The encoded features from the environment, robot, and configuration space are fused and processed by decoder modules to predict the next motion step. The dimensions of key features are indicated in brackets.}
    \label{fig:planner_pipeline}
\end{figure*}

\section{Problem Definition}
This section formally defines the neural motion planning problem for serial robotic manipulators. We consider an $n$ degree-of-freedom robotic arm comprising $m$ physical links connected by revolute joints, where the joint configuration is represented as a vector $\mathbf{q} = [q_1, q_2, \dots, q_n]^\top \in \mathbb{R}^n$ with each component corresponding to a joint angle.
The neural motion planning task involves learning a policy $\pi_{\theta}$ that can generate collision-free trajectories based on some representation of the environment and robot. These representations encode environmental geometry and robot-specific information in various forms, which range from point clouds and voxel grids to implicit neural representations such as SDF. Different approaches may emphasize different aspects of this representation, with some focusing primarily on environmental encoding while others incorporate explicit robot geometry modeling.
At each timestep $t$, the policy processes the current joint state $\mathbf{q}_t$, the target joint state $\mathbf{q}_g$, and features extracted from the chosen representations. Based on this input, it generates the subsequent joint state $\mathbf{q}_{t+1}$, iteratively guiding the manipulator toward $\mathbf{q}_g$ while avoiding collisions until the target is reached within a specified tolerance. The key challenge lies in effectively designing and integrating these representations to enable robust collision-aware motion generation across diverse environments and robot configurations.

\section{Methodology}

In this work, we propose SAMP, a neural motion planning method that takes as input the robot’s current and goal joint configurations, along with a fused spatial anchor feature jointly encoding environmental structure and robot geometry, to generate collision-free motions in an end-to-end manner. An overview of our SAMP pipeline is provided in Fig.~\ref{fig:planner_pipeline}. The spatial anchor feature is built by aligning two complementary representations: environmental features from anchor points with precomputed SDF values (Section \ref{sec:env_rep}), and robot geometry features from sampled points around links with inferred neural SDF values (Section \ref{sec:rob_pre}). These features are spatially aligned via a joint grid sampling strategy (Section \ref{sec:feature_align}) and fused with encoded joint configurations to produce kinematically feasible motions that implicitly satisfy environmental constraints. The network architecture and training procedure are described in detail in Section~\ref{sec:network}. 

\subsection{Spatial Anchor Feature}\label{sec:saf}

This section elaborates on the construction process of the spatial anchor feature, including the environmental representation based on anchor points and the robot geometry encoding using sampled points of arm links.

\subsubsection{Environmental Feature}\label{sec:env_rep}  
 
To represent the spatial structure of the robot’s workspace, we construct an environmental SDF grid \(\mathcal{F}^\text{env} \in \mathbb{R}^{n_e \times n_e \times n_e}\), where \(n_e\) denotes the grid resolution along each spatial dimension. This volumetric representation discretizes the workspace into a 3D grid of signed distance values. A set of anchor points \(\mathcal{A} = \{\mathbf{p}_a \mid \mathbf{p}_a \in \mathbb{R}^{n_e \times n_e \times n_e}\}\) is uniformly distributed within this grid, with each point corresponding to a discrete location in the reachable workspace. These anchor points provide dense and comprehensive spatial coverage of the environment. Each anchor point queries signed distance values $d^e$ directly using \(f_\theta^e\) to capture obstacle geometry. Unlike previous work that relies on occupancy representations, SAMP encodes environmental collision constraints through the obstacle distance feature \(\mathcal{F}^\text{env} = \{d^e_{a}\}_{a\in\mathcal{A}}\), which aggregates continuous SDF values across all anchor points.

Our framework flexibly incorporates environmental SDF features and is agnostic to the specific representation of SDF. It supports both neural implicit representations \cite{iSDF2022}, \cite{shine} and traditional voxel-based methods \cite{voxblox}, \cite{voxfield}. Rather than focusing on building novel SDF construction methodologies, we leverage neural-based SDF representations for environments constructed through established pipelines \cite{iSDF2022}, which can be either precomputed offline or updated online according to the application context and available resources.

\subsubsection{Robot Geometry Representation}\label{sec:rob_pre}

Our method for robot geometry encoding begins with sampling a set of points around each link surface. For each link $l_i$ ($i = 1,2,\dots,m$), we generate points within a specified distance, forming the set $\mathcal{P}^{\text{robot}}_i = \{\mathbf{p}_{i,k} \in \mathbb{R}^3 \mid k=1,2,\dots,M_i\}.$ The complete robot point set is then defined as $\mathcal{P}^{\text{robot}} = \bigcup_{i=1}^m \mathcal{P}^{\text{robot}}_i$, and the number of all robot points is $ M = \sum_{i=1}^m M_i$. These points are processed through the proposed neural SDF model, as detailed in Section \ref{sec:feature_align}, to predict their signed distances relative to the robot body under various joint configurations. This design offers two key advantages for training the neural policy for collision avoidance. First, by sampling beyond the immediate surface, the model encounters a wider range of potential collision scenarios during training. This strategy enhances the network's awareness of proximity and potential collisions during training. By contrast, surface-only sampling provides limited near-collision exposure when tracking expert trajectories, offering weaker learning signals for collision avoidance. Second, the SDF information of the sampled points enables adaptive loss weighting based on distance to the robot surface, allowing more informed gradient updates during optimization. The resulting robot geometry feature with SDF information, \(\mathcal{F}^\text{robot} = \{d^r_{k}\}_{k\in\mathcal{P}^\text{robot}}\), captures the manipulator's spatial distribution.

\begin{figure}[!tp]
    \centering
    \includegraphics[width=1.0\linewidth]{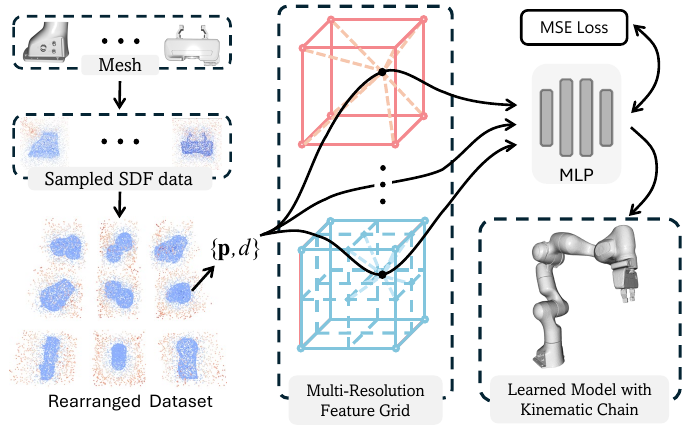}
    \caption{
The training pipeline learns the robot’s SDF through multi-resolution feature grids, which capture geometric structures at different hierarchical scales. An MLP is then used to decode continuous SDF values from these feature representations.
    }
    \label{fig:sdf_pipeline}
\end{figure}

To enable more efficient and accurate inference of the robot's SDF, we propose a unified training approach for the neural SDF model, and the learning procedure is summarized in Fig. \ref{fig:sdf_pipeline}. Inspired by \cite{RDF}, we construct the training dataset by first normalizing each high-fidelity mesh $\mathcal{M}_i$ of link $l_i$ to fit within an axis-aligned bounding box of $[-1,1]^3$ meters, and then uniformly sampling $N_i$ points $\{\mathbf{p}_{i,j} \in \mathbb{R}^3 \mid j =1,2,\dotsc,N_i\}$ within this normalized volume. For each sampled point $\mathbf{p}_{i,j}$, the SDF value $d_{i,j}$ is computed as the signed distance to the belonging link's mesh surface, with positive values indicating points outside the mesh and negative values representing points inside the mesh. The resulting training dataset for the $i$-th link is formally defined as $\mathcal{D}_i = \{ (\mathbf{p}_{i,j}, d_{i,j}) \mid j =1,2,\dotsc,N_i \}$, where $N_i$ is the total number of sampled points for link $l_i$.

Unlike prior works that typically train individual neural networks for each robot link \cite{RDF},\cite{percep_local}, we adopt a unified learning procedure that enables simultaneous learning of SDF for all links within a shared model. Specifically, we aggregate the datasets for all links by spatially offsetting each link's point cloud with a predefined translation \(\mathbf{t}_i\), such that their occupied regions in the unified domain do not overlap, as illustrated in Fig. \ref{fig:sdf_pipeline}. This allows the model to treat the collection of links as a single composite scene and to jointly learn their implicit representations consistently. Formally, the relocated dataset for the \(i\)-th link is given by:
\begin{equation}
\mathcal{D}_i^{'} = \{\mathbf{p}_i' = \mathbf{p}_i + \mathbf{t}_i \mid \mathbf{p}_i \in \mathcal{D}_i,\, \mathbf{t}_i \in \mathbb{R}^3\},
\end{equation}
and the complete aggregated training dataset is defined as:
\begin{equation}
\mathcal{D}_\text{all} = \bigcup_{i=1}^{L} \mathcal{D}_i^{'}.
\end{equation}

For the network architecture, we use multiresolution hash encoder maps 3D coordinates to high-dimensional feature vectors across \(L\) distinct resolution levels building upon Instant-NGP \cite{mueller2022instant}. At each level \(l\), the embedding feature \(\phi_l(\mathbf{p})\) is obtained by hashing the grid vertices and applying trilinear interpolation. The multi-scale features from all levels are concatenated and subsequently passed through an MLP to predict the SDF value:
\begin{equation}
d^\text{robot} = f_{\theta}\left(\bigoplus_{l=1}^L \phi_l({\mathbf{p}})\right),
\end{equation}
where \(\bigoplus\) denotes feature concatenation across all levels, and \(f_{\theta}(\cdot)\) represents the MLP-based decoder that outputs the predicted SDF value. The model is trained by minimizing the mean squared error (MSE) loss. For brevity, we use $f_{\theta}^r$ to represent the trained robot SDF model. The complete training procedure is summarized in Algorithm~\ref{alg:unified_robot_sdf}.

\begin{algorithm}[tp!]
\caption{Unified Robot SDF Learning}
\label{alg:unified_robot_sdf}
\KwIn{Link meshes $\{\mathcal{M}_i\}$ for all robot links}
\KwOut{Trained neural SDF model $f_\theta^r$}

\SetKwFunction{ConstructDataset}{ConstructDataset}
\SetKwFunction{RearrangeDataset}{RearrangeDataset}
\SetKwFunction{TrainNeuralSDF}{TrainNeuralSDF}

$\{\mathcal{D}_i\} \leftarrow$ \ConstructDataset{$\{\mathcal{M}_i\}$}\;
$\mathcal{D}_\text{all} \leftarrow$ \RearrangeDataset{$\{\mathcal{D}_i\}$}\;
$f_{\theta}^r \leftarrow$ \TrainNeuralSDF{$\mathcal{D}_\text{all}$}\;

\SetAlgoNoLine
\SetKwProg{Fn}{Function}{:}{}
\Fn{\ConstructDataset{$\{\mathcal{M}_i\}$}}{
    \For {each link $i$}{
        Normalize mesh $\mathcal{M}_i$ to the volume $[-1, 1]^3$\;
        Sample points $\mathbf{p}_{i,j}$ in the normalized volume\;
        Compute signed distance $d_{i,j}$ for each $\mathbf{p}_{i,j}$\; 
        $\mathcal{D}_i = \{(\mathbf{p}_i, d_i)\}$\;
    }
    \Return $\{\mathcal{D}_i\}$\;
}

\SetAlgoNoLine
\SetKwProg{Fn}{Function}{:}{}
\Fn{\RearrangeDataset{$\{\mathcal{D}_i\}$}}{
    \For {each link $i$}{
        $\mathcal{D}_i' = \{(\mathbf{p}_i', d_i) \mid \mathbf{p}_i' = \mathbf{p}_i + \mathbf{t}_i\}$\;
    }
    $\mathcal{D}_\text{all} = \bigcup_i \mathcal{D}_i'$\;
    \Return $\mathcal{D}_\text{all}$\;
}

\SetAlgoNoLine
\SetKwProg{Fn}{Function}{:}{}
\Fn{\TrainNeuralSDF{$\mathcal{D}_\text{all}$}}{
    \While{not reach max epoch}{
        Sample minibatch $(\mathbf{p}, d)$ from $\mathcal{D}_\text{all}$\;
       $\Phi(\mathbf{p}) = \bigoplus_{l=1}^L \phi_l(\mathbf{p})$\;
        $\hat{d} = f_\theta(\Phi(\mathbf{p}))$\;
        Optimize model by minimizing MSE loss $\sum (\hat{d} - d)^2$\;
    }
    \Return Trained SDF model $f_{\theta}^r$
}
\end{algorithm}

\subsubsection{Feature Alignment}\label{sec:feature_align}

\begin{figure}[tp]
    \centering
    \includegraphics[width=0.78\linewidth]{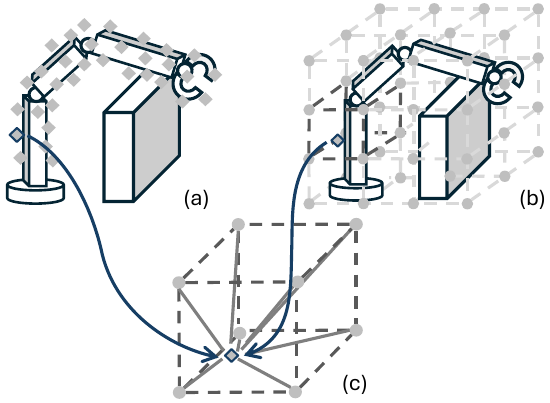}
    \caption{
    Visualization of spatial anchor feature extraction and alignment in the SAMP framework. (a) Dense points (represented by grey diamond points) are sampled around the robot links within the workspace, representing the geometric structure of robots. (b) Predefined anchor points (represented by grey circular points) are distributed throughout the environment, with each point encoding signed distance values to the nearest obstacles, forming the environmental spatial representation. (c) Alignment process through joint grid sampling that fuses both robot geometry features and environmental obstacle features into a unified representation for neural motion planning. 
    }
    \label{fig:sdf_align}
\end{figure}

Motivated by \cite{grid_sampler}, given the environmental obstacle feature $\mathcal{F}^\text{env}$ and the robot geometry feature $\mathcal{F}^{\text{robot}}$, we employ joint grid sampling to extract environmental features at each robot point location, as shown in Fig. \ref{fig:sdf_align}. For each robot point $\mathbf{p}_k = p_{k,D}$, we normalize its coordinates to the range $[0, 1]$ to align with the environmental feature coordinate system and calculate the corresponding index $I_{k,D}$:
\begin{equation}
I_{k,D} = \frac{p_{k,D} - p_{\text{min},D}}{p_{\text{max},D} - p_{\text{min},D}} \cdot (n_e -1), \quad D \in \{x,y,z\},
\end{equation}
where ${p}_{\text{min}, D}$ and ${p}_{\text{max}, D}$ define the workspace boundaries along the axis $D$. 

For obtaining the environment feature at $\mathbf{p}_k$, we let $\lfloor I_{k,D} \rfloor$ and $\lfloor I_{k,D} \rfloor+1$ denote the two neighboring grid indices, and define the fractional coordinates:
\begin{equation}
u = I_{k,x} - \lfloor I_{k,x} \rfloor, \quad 
v = I_{k,y} - \lfloor I_{k,y} \rfloor, \quad 
w = I_{k,z} - \lfloor I_{k,z} \rfloor.
\end{equation}
Then, the environmental feature is obtained via trilinear interpolation:
\begin{equation}
\begin{aligned}
\mathcal{G}(\mathcal{F}^\text{env}, \mathbf{p}_k) &= \sum_{i=0}^{1}\sum_{j=0}^{1}\sum_{m=0}^{1} w_{ijm} \cdot \\
&\quad \mathcal{F}^\text{env}(\lfloor I_{k,x} \rfloor + i, \lfloor I_{k,y} \rfloor + j, \lfloor I_{k,z} \rfloor + m),
\end{aligned}
\end{equation}
where $i,j,m \in \{0,1\}$ denote the binary offsets toward the neighboring grid vertices along each axis, and the interpolation weights are defined as
\begin{equation}
\begin{aligned}
w_{ijm}(u,v,w) &= (i \cdot u + (1-i)(1-u)) \cdot \\
               &\quad (j \cdot v + (1-j)(1-v)) \cdot \\
               &\quad (m \cdot w + (1-m)(1-w)).
\end{aligned}
\end{equation}
This operation is applied independently to each feature channel $C$, and the interpolated feature for all transformed robot points is $\mathcal{G}' \in \mathbb{R}^{M \times C}$.

\subsection{Network Architecture and Training Details}\label{sec:network}

Our method trains the proposed SAMP using supervised learning on a dataset $\mathcal{D}^\text{traj}$ of expert demonstration trajectories with random valid start and goal configurations. Each trajectory segment ${\mathbf{q}_{t}, \mathbf{q}_{t+1}, \dots, \mathbf{q}_{t+n}}$ in $\mathcal{D}^\text{traj}$ is generated by an established planner \cite{cuRobo}, from which we extract input-output pairs $(\mathbf{q}_t, \mathbf{q}_g, \mathcal{A}, \mathcal{P}^\text{robot}) \rightarrow \mathbf{q}_{t+1}$ for training.

\subsubsection{Network Architecture}

\begin{figure}[tp]
    \centering
    \includegraphics[width=0.9\linewidth]{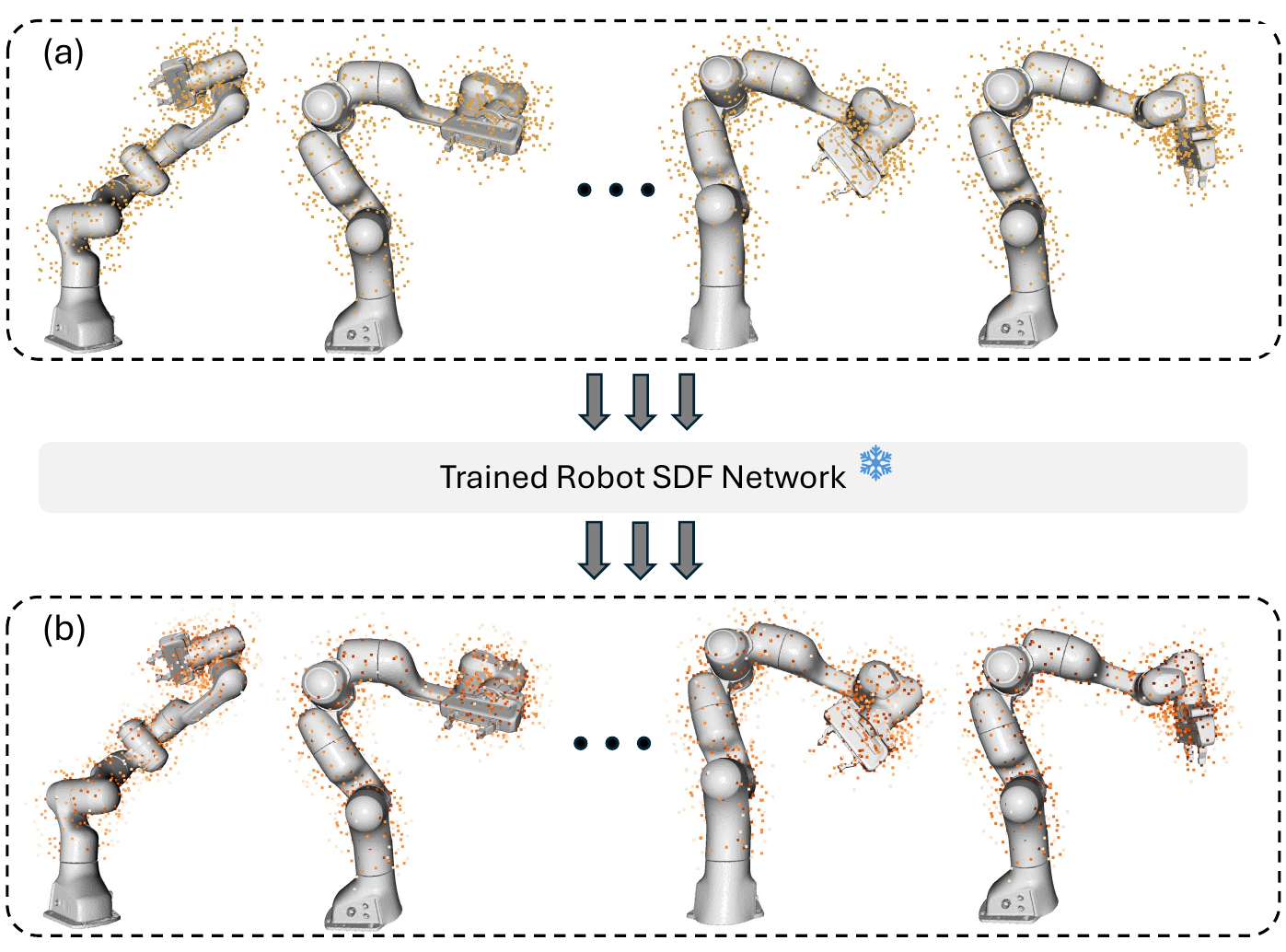}
    \caption{Illustration of SDF inference for sampled robot points. (a) For different robot configurations, the pre-sampled points around the link surfaces and transformed into their corresponding positions in workspace coordinates. (b) These points are then processed by the trained robot SDF network to obtain their signed distances to the nearest robot surface, where darker-colored points indicate closer proximity to the surface.}
    \label{fig:sample_point}
\end{figure}

Given the current and goal robot joint configurations, \(\mathbf{q}_t, \mathbf{q}_g \in \mathbb{R}^n\), and the environmental spatial anchor grid \(\mathcal{A} \in \mathbb{R}^{n_e \times n_e \times n_e}\), our model encodes a latent representation for motion prediction as follows. First, the environmental SDF grid \(\mathcal{F}^\text{env}\) is encoded using a 3D convolutional encoder \(\Phi_{\text{3D}}\), producing spatially-aware feature volumes. To capture obstacle information at relevant locations, we use a joint grid sampling operation at the positions of the robot points. For robot points \(\mathcal{P}^{\text{robot}}\) sampled on each relevant link, we apply a transformation operator \(T_{\mathbf{q}}(\cdot)\), which maps the sampled points to their correct positions in the workspace according to the given joint configuration \(\mathbf{q}\). The final environmental SDF feature is defined as:
\begin{equation}
    \mathcal{G}' = \mathcal{G}\left(\Phi_{\text{3D}}(\mathcal{F}^\text{env}), T_{\mathbf{q_t}}(\mathcal{P}^{\text{robot}}) \right).
\end{equation}
In parallel, the robot point cloud is processed by a pretrained neural SDF model for the robot, as illustrated in Fig. \ref{fig:sample_point}, producing robot SDF features \(\mathcal{F}^\text{robot} = f_\theta^r(T_{\mathbf{q_t}}(\mathcal{P}^{\text{robot}}))\). Additionally, a point cloud encoder $\Phi_{pe}$ is utilized to extract structural features from the sampled points of the robot \cite{point_net}.

For joint configuration encoding, the current and goal configurations are concatenated and encoded by an MLP to obtain a configuration space feature:
\begin{equation}
\mathcal{F}^{\text{c}} = \text{MLP}^\text{I}([\mathbf{q}_t, \mathbf{q}_g]).
\end{equation}
Then, the environmental anchor features \(\mathcal{G}'\) and robot SDF features \(\mathcal{F}^\text{robot}\) form a unified spatial latent representation:
\begin{equation}
\mathcal{F}^{s} = \phi(\mathcal{G}', \mathcal{F}^\text{robot}),
\end{equation}
where \(\phi\) denotes the feature fusion operation, implemented via feature concatenation and a subsequent decoder \text{MLP}$^\text{II}$. Finally, the fused feature:
\begin{equation}
    \mathcal{F}^{'}=(\mathcal{F}^c, \mathcal{F}^s,\Phi_{pe}(T_{\mathbf{q_t}}(\mathcal{P}^{\text{robot}})),
\end{equation} 
which is fed into a decoder network to predict the change in joint configuration for the next step:
\begin{equation}
\Delta \mathbf{q}_{t+1} = \text{MLP}^\text{III}(\mathcal{F}^{'}).
\end{equation}
The final predicted next configuration is obtained by:
\begin{equation}
\hat{\mathbf{q}}_{t+1} = \mathbf{q}_t + \Delta\mathbf{q}_{t+1}.
\end{equation}

\subsubsection{Loss Function}  
The training objective consists of several loss terms, including point matching accuracy, collision avoidance, and expert supervision of joint configurations. 

\textbf{Point matching loss:} Inspired by \cite{motionpolicynetwork}, the point matching loss \(\mathcal{L}_{\text{match}}\) enforces geometric consistency between the robot point clouds corresponding to the predicted configuration \(\hat{\mathbf{q}}_{t+1}\) and the supervised configuration \(\mathbf{q}_{t+1}\). By directly minimizing the distance between the transformed robot point clouds under the two configurations, this loss encourages the predicted joint configuration to result in a robot pose that closely matches the supervised target pose in 3D space. Then, the point matching loss is defined as follows:
\begin{equation}
\begin{split}
\mathcal{L}_{\text{match}} =\ & \left\| T_{\hat{\mathbf{q}}_{t+1}}(\mathcal{P}^{\text{robot}}) - T_{\mathbf{q}_{t+1}}(\mathcal{P}^{\text{robot}}) \right\|_1 \\
& + \left\| T_{\hat{\mathbf{q}}_{t+1}}(\mathcal{P}^{\text{robot}}) - T_{\mathbf{q}_{t+1}}(\mathcal{P}^{\text{robot}}) \right\|_2\ ,\end{split}
\end{equation}
where \(\|\cdot\|_1\) denotes the element-wise L1 norm and \(\|\cdot\|_2\) denotes the squared L2 norm.

\begin{algorithm}[t]
\caption{SAMP}
\label{alg:samp}
\KwIn{
  Dataset $\mathcal{D}^\text{traj}$, current state $\mathbf{q}_t$, goal $\mathbf{q}_g$, anchor points $\mathcal{A}$, robot points $\mathcal{P}^\text{robot}$, environment SDF function $f_\theta^e$, robot SDF model $f_\theta^r$
}
\KwOut{
 Learned policy model $\pi_\theta$; predicted action $\mathbf{u}_{t+1}$
}

\tcp*[h]{\textbf{Training phase}} \\
\Indp\For {each iteration}{
 Sample minibatch $\{(\mathbf{q}_t, \mathbf{q}_g, \mathbf{q}_{t+1}^*)\}$ from $\mathcal{D}^\text{traj}$\;
 $\mathcal{P}_{\mathbf{q}_t}^\text{robot} \gets T_{\mathbf{q_t}}(\mathcal{P}^{\text{robot}})$\;
 $\mathcal{F}^\text{robot} \gets f_\theta^r(\mathcal{P}_{\mathbf{q}_t}^\text{robot})$\;
 $\mathcal{F}^\text{env} \gets f_\theta^e(\mathcal{A})$\;
 $\mathcal{G}' \gets \mathcal{G}(\Phi_{\text{3D}}(\mathcal{F}^\text{env}), T_{\mathbf{q_t}}(\mathcal{P}^{\text{robot}}) )$\;
 $\Delta\mathbf{q}_{t+1} \gets \pi_\theta([\mathbf{q}_t, \mathbf{q}_g], [\mathcal{G}', \mathcal{F}^\text{robot}], \Phi_{pe}(T_{\mathbf{q_t}}(\mathcal{P}^{\text{robot}})))$\;
 $\mathcal{L} \gets \mathcal{L}_\text{total}(\mathbf{q}_t + \Delta\mathbf{q}_{t+1}, \mathbf{q}_{t+1}^*)$\;
 Update $\pi_\theta$ to minimize $\mathcal{L}$\;
}
\Return $\pi_\theta$ \;
\Indm
\vspace{1mm}
\tcp*[h]{\textbf{Inference phase}} \\
\Indp\While{not reach goal}{
 $\Delta\mathbf{q}_{t+1} \gets \pi_\theta([\mathbf{q}_t, \mathbf{q}_g], [\mathcal{G}', \mathcal{F}^\text{robot}], \Phi_{pe}(T_{\mathbf{q_t}}(\mathcal{P}^{\text{robot}})))$\;
 $\mathbf{u}_{t+1} \gets \mathbf{q}_t + \Delta\mathbf{q}_{t+1}$\;
 $\mathbf{q}_t \gets \mathbf{u}_{t+1}$\;
}
\Indm\end{algorithm}

\textbf{Collision loss:} 
The collision avoidance loss penalizes robot joint configurations predicted by the network that would cause sample points on the robot to collide with obstacles. We use grid sampling to align the SDF of spatial anchor points with the robot's sample points, estimating the collision probability for each point. Penalty weights are assigned based on the distance from each sample point to the robot surface. This approach provides denser loss feedback and makes the robot less sensitive to minor constraint violations. The collision loss is defined as:
\begin{equation}
\mathcal{L}_{\text{coll}} = \frac{1}{M} \sum_{i=1}^{M} w_i \cdot \max(0, d_{\text{margin}} - C_i),
\end{equation}
where \( C_i= \mathcal{G}(\mathcal{F}^\text{env}, T_{\mathbf{q_t}}(\mathcal{P}^{\text{robot}}))\) denotes SDF value of the obstacles at the \(i\)-th robot sample point and 
and \( w_i = \exp(- (d_i - d_{\text{target}})) \), where \( d_i \) is the distance from the sample point to the robot surface and \( d_{\text{target}} \) is a target margin hyperparameter.

\textbf{Configuration loss:}
This loss directly penalizes the difference between the predicted joint configuration \(\mathbf{q}^{\text{pred}}\) and the target configuration \(\mathbf{q}^{\text{target}}\) using the squared L2 norm. By minimizing this loss, the network is encouraged to output joint values that are as close as possible to the supervised targets, which helps to improve the accuracy of joint prediction and stabilize the learning process:
\begin{equation}
\mathcal{L}_{\text{config}} = \|\hat{\mathbf{q}}_{t+1} - \mathbf{q}_{t+1}\|_2^2.
\end{equation}

The complete training objective combines these components, shown as follows:
\begin{equation}
\mathcal{L}_{\text{total}} = \lambda_1\mathcal{L}_{\text{match}} + \lambda_2\mathcal{L}_{\text{coll}} + \lambda_3\mathcal{L}_{\text{config}},
\end{equation}
where the coefficients \(\lambda_1\), \(\lambda_2\), and \(\lambda_3\) serve as weighting factors that balance the relative contributions of each loss term in the overall objective. The complete training pipeline is presented in Algorithm~\ref{alg:samp}.

\begin{table}[tbp]
\centering
\caption{Key parameters and their descriptions}
\label{tab:parameters}
\begin{tabular}{>{\centering\arraybackslash}p{0.8cm} >{\centering\arraybackslash}p{1.0cm} p{5.5cm}}
\toprule
\textbf{Param} & \textbf{Value} & \textbf{Description} \\
\midrule
$n$      & 7        & Degrees of freedom of the manipulator \\
$m$      & 8        & Number of physical links of the manipulator \\
$n_e$    & 160      & Resolution of the anchor points \\
$M$      & 1024     & Number of sampled robot points \\
$N$      & $5 \times 10^5$ & Number of sampled points for training SDF \\
$t$      & 3      & Offset for generating rearrangement dataset \\
$L$      & 32       & Number of feature levels in the hash encoder \\
$C$      & 4        & Number of feature channels in 3D convolution \\
$d_{\text{margin}}$ & 0.09 m & Safety margin used in collision loss \\
$d_{\text{target}}$ & 0.015 m & Activation threshold for sampled point distances \\
\bottomrule
\end{tabular}
\end{table}

\begin{figure}
    \centering
    \includegraphics[width=1.0\linewidth]{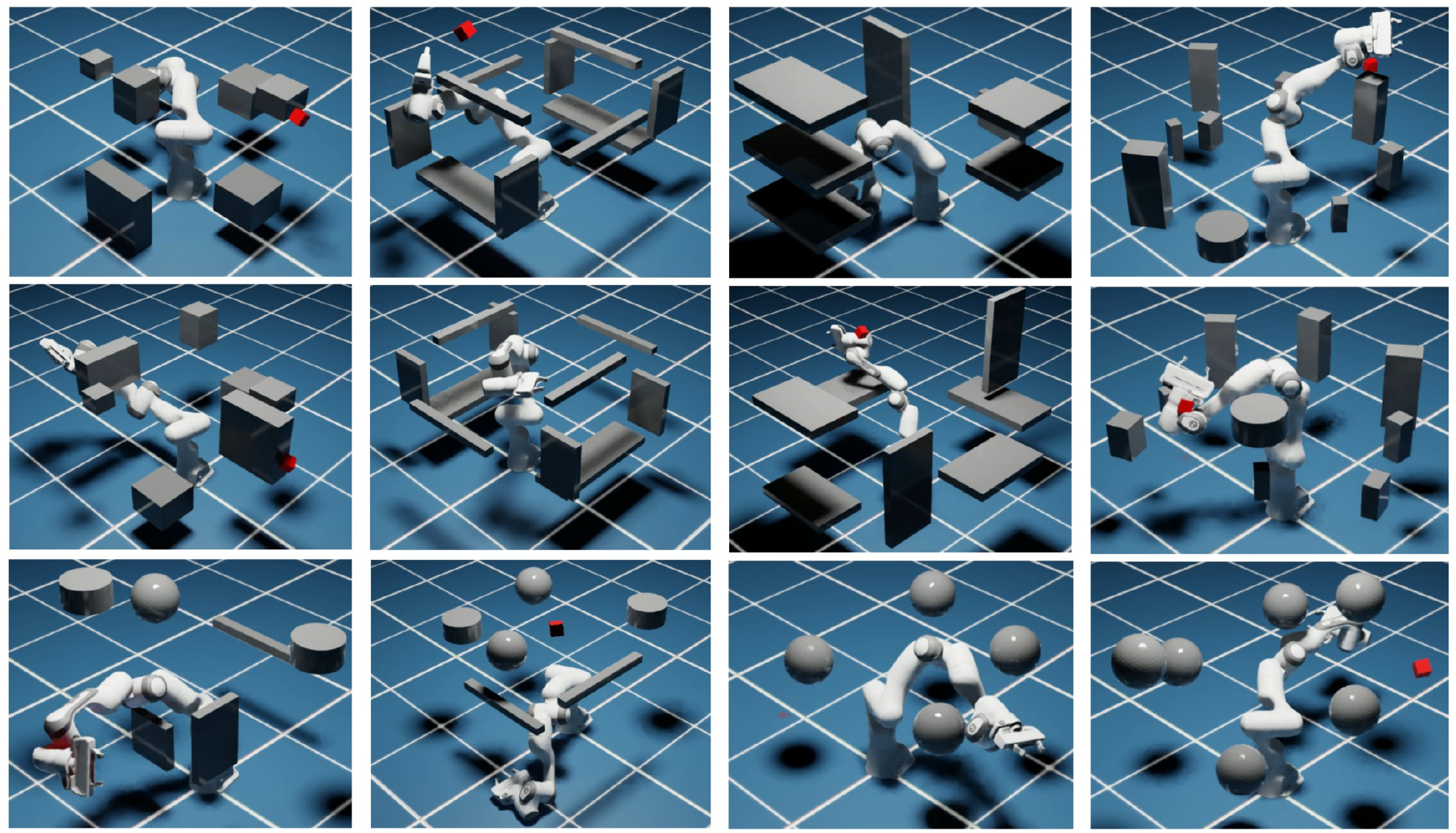}
    \caption{Examples of simulated environments used for collision-free trajectories collection. The panda robot is placed in diverse cluttered scenes with randomly generated obstacles of varying primitive shapes (boxes, cylinders, and spheres) and spatial arrangements. Each scene is paired with a designated target object (highlighted in red).}
    \label{fig:exp_environ}
\end{figure}

\section{Experiments}

\subsection{Experiment Setup}

\subsubsection{Implementation Details}
Our implementation starts with data collection in simulation using Isaac sim, where we construct 40 environments in total, as depicted in Fig.~\ref{fig:exp_environ}. Specifically, 32 environments are allocated for training purposes and 8 for testing to evaluate the method's performance. For training, we generate 1,000 expert trajectories per environment using an established planner~\cite{cuRobo}, with each trajectory comprising 50 planning steps, yielding 1.6 million samples in total. For evaluation, 200 additional expert trajectories are collected, covering both training and held-out environments, enabling assessment on both seen and unseen scenarios. All experiments are conducted on an NVIDIA RTX 3080Ti GPU. The model is trained for 160 epochs with the initial learning rate 0.0001, and the batch size is set to 50. Furthermore, Table~\ref{tab:parameters} summarizes key hyperparameters for reference.

\subsubsection{Baselines and Metrics} 
We conduct our method against both classical optimization-based motion planning methods and recent learning-based approaches. 
Specifically, cuRobo~\cite{cuRobo}, with both optimization and sampling variants, serves as the model-based baseline. MPNet~\cite{MPN}, SIMPNet~\cite{simpnet}, and MPiNet~\cite{motionpolicynetwork} represent learning-based motion planning methods that leverage neural networks for motion prediction. All methods are evaluated using four metrics: Solution Time (Soln. Time), which reflects the average computation time per trajectory; Path Length, measuring the trajectory distance of the end-effector; Success Rate, denoting the percentage of successful trials; Reach Rate indicating the proportion of trajectories that successfully reach the goal regardless of collisions and Collision Rate (Coll. Rate), representing the percentage of trajectories that result in collisions. Results are averaged over all trials in the respective datasets, with standard deviations reported to ensure fair comparison.

\begin{figure*}[t]
    \centering
    \includegraphics[width=0.95\linewidth]{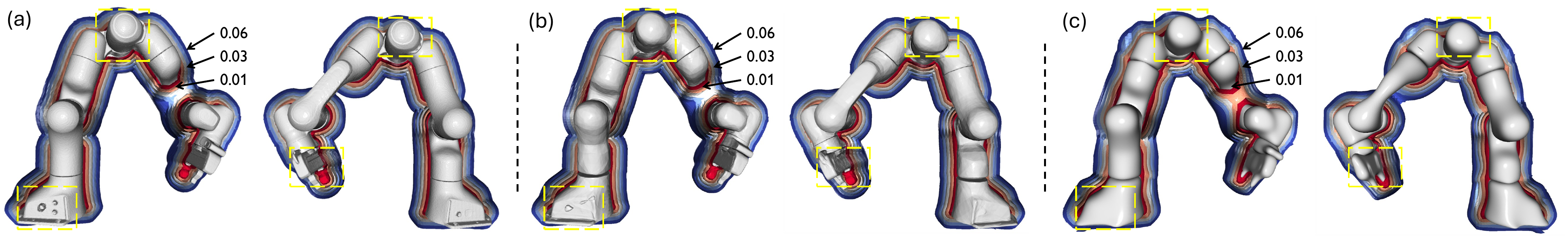}
    \caption{Comparison of robot SDF representations across different neural-based methods: (a) Our approach, (b) NN-SDF, and (c) RDF\cite{RDF}. Each subfigure shows offset layers from 0.01 m to 0.06 m around the robot links, visualized as distinct colored bands. Representative regions are highlighted with orange boxes for detailed comparison.}
    \label{fig:vis_robot_sdf}
\end{figure*}

\subsection{Evaluation of Robot Geometry Learning}
We first present the evaluation results of robot SDF learning accuracy achieved by our proposed method. Our approach is compared against several baselines, including neural based SDF (NN-SDF) \cite{nn_sdf} implemented with an MLP and robot distance field (RDF) \cite{RDF}. As shown in Table~\ref{tab:aba_study}, our method consistently outperforms all baselines across all evaluated metrics. Specifically, our approach achieves the lowest mean absolute error (MAE) and root mean square error (RMSE) both in the overall test set and in the \texttt{Near} and \texttt{Far} regions. Notably, our method attains an MAE of $3.30$\,mm and an RMSE of $6.81$\,mm, significantly improving upon the best baseline methods. In addition, our approach demonstrates the fastest training time with less than $90$\,s, achieving more than an order of magnitude speed-up compared to others, while maintaining competitive inference speed. These results clearly highlight the superior accuracy and efficiency of our method for learning SDF.

To qualitatively evaluate reconstruction performance, Fig.~\ref{fig:vis_robot_sdf} shows a comparison of the reconstructed side surfaces obtained using different methods. The figure compares sampled offset surfaces produced by our method, NN-SDF, and RDF at distances for $0.01$\,m, to $0.06$\,m from the robot links. Our approach generates smooth and uniformly distributed offset layers that closely conform to the robot geometry, ensuring consistent coverage around both cylindrical and joint regions. In contrast, NN-SDF exhibits noticeable distortions and irregularities near areas of high curvature, while RDF produces discontinuous or overly diffused offset regions. The highlighted yellow boxes mark representative areas where these differences are most pronounced, further demonstrating the improved accuracy and robustness of our method in capturing the robot's geometry.


\begin{table}[tbp]
\setlength{\abovecaptionskip}{0pt} 
\setlength{\belowcaptionskip}{0pt} 
\fontsize{7}{8}\selectfont
\renewcommand\arraystretch{1.2}
\setlength\tabcolsep{3pt} 
\centering
\begin{threeparttable}
\caption{Comparison of SDF accuracy and computational efficiency across different learning-based methods}\label{tab:aba_study}
\begin{tabular}{cccccccccc}
\toprule[0.3mm]
\multirow{2}{*}{\makecell[c]{Methods}} & \multicolumn{3}{c}{MAE (mm)} &  \multicolumn{3}{c}{RMSE (mm)}  & \multirow{2}{*}{\makecell[c]{Training\\Time (s)}} & \multirow{2}{*}{\makecell[c]{Inference\\Time (ms)}} \\ 
\cline{2-7}
& All & Near & Far & All & Near & Far &  \\
\midrule
NN-SDF                 & 3.87 & 1.33 & 4.76 & 7.11  & 3.54 & 7.97  & 690    & 5.22 \\
RDF (N=8)          & 4.70 & 2.95 & 5.32 & 7.52  & 4.85 & 8.24  & 870    & \textbf{1.88} \\
RDF (N=24)         & 3.59 & 1.2 & 4.42 & 6.99  & 3.55 & 7.82  & 14658  & 14.25   \\  
Ours  & \textbf{3.30} & \textbf{0.72} & \textbf{4.22} & \textbf{6.81}  & \textbf{3.26} & \textbf{7.65}  & \textbf{80}     & 2.58 \\  
\bottomrule[0.3mm]
\end{tabular}
\begin{tablenotes}[para,flushleft] 
\footnotesize
Note: The best is emphasized in bold. \texttt{Near} refers to test SDF values that are less than 0.05\,m, while \texttt{Far} describes distances ranging from 0.05 to 0.15\,m.
\end{tablenotes}
\end{threeparttable}
\end{table}

\begin{figure}[tp!]
    \centering
    \includegraphics[width=0.9\linewidth]{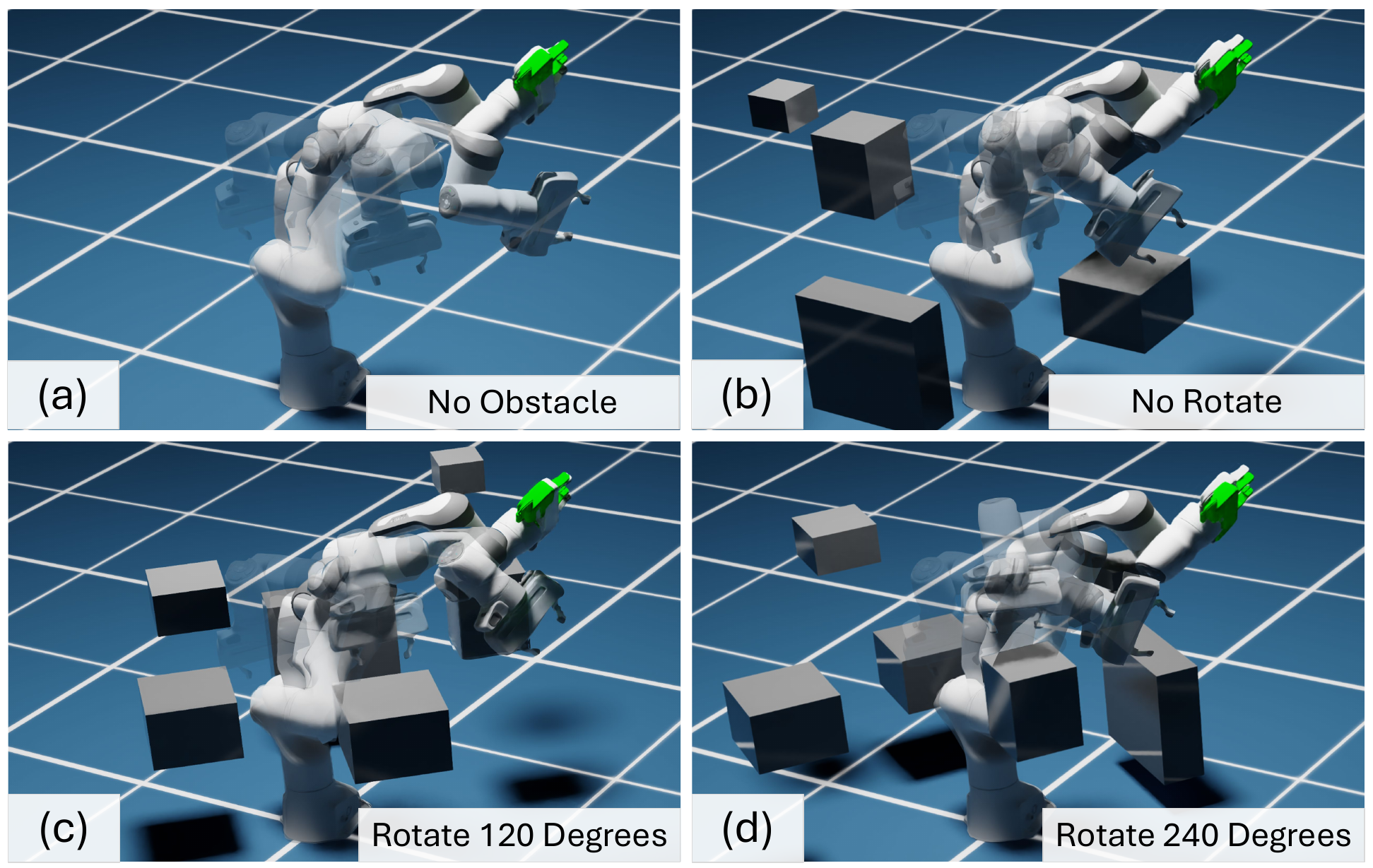}
    \caption{Trajectory comparison under global rotations of the obstacle set with fixed internal structure. Each subfigure (a)–(d) demonstrates the motion trajectory generated from identical start and goal poses, showing the policy's robustness to the changes of different obstacle arrangements.}
    \label{fig:traj_diff_obs}
\end{figure}

\subsection{Evaluation of Motion Policy}

Table~\ref{tab:samp_comp} presents a quantitative comparison of motion planning performance across various state-of-the-art methods. CuRobo (Opti.) serves as the expert planner used to generate the trajectories for policy training with optimal results across all metrics. Compared to other SOTA neural motion planners, our proposed SAMP method achieves a substantial improvement in both success rate and collision rate in both training and testing environments. Specifically, SAMP (160 anchors) attains a success rate of $90$\,\% in the training environment and $81$\,\% in the testing environment, while reducing the collision rate to $6$\,\% and $7$\,\%, markedly outperforming other SOTA methods such as MPNet \cite{MPN} and SIMPNet \cite{simpnet}. 

In terms of efficiency, SAMP achieves the shortest solution time (below $0.09$\,s), enabling near real-time trajectory generation. This advantage primarily comes from its end-to-end design, where the entire trajectory is predicted in a single forward pass. In contrast, other neural-based baselines such as MPNet, SIMPNet, and MPiNet typically require iterative waypoint generation and stepwise collision checking, or employ deeper network architectures with higher computational cost per inference. As a result, although they can also leverage neural predictions, their execution latency is higher than SAMP. The combination of compact architecture design and robot-centric free-space representation allows SAMP to produce collision-aware trajectories both faster and with higher success rates.

To further evaluate the adaptability of our planning policy, we conduct experiments under varying obstacle conditions, as illustrated in Fig. \ref{fig:traj_diff_obs}. Initially, the robot operates in an obstacle-free environment, where the end-effector moves directly from the start to the goal position without any detours (Fig. \ref{fig:traj_diff_obs}(a)). Subsequently, we introduce obstacles into the workspace and systematically apply different rigid orientations to the same obstacle arrangement. As shown in Fig. \ref{fig:traj_diff_obs}(b)–(d), our policy automatically generates distinct, collision-free trajectories in response to the altered obstacle configurations. These results demonstrate that the policy is capable of perceiving and reacting to changes in the environment, producing effective avoidance maneuvers while reliably reaching the goal.

\begin{table*}[tbp]
\centering\caption{Comparison of motion planning performance across traditional methods (cuRobo opti. and Sampling), neural-based methods (MPNet, SIMPNet, MPiNet), and our proposed approach with different anchor resolutions (100 and 160). Metrics are reported for both training and testing environments, including solution time, path length, success rate, and collision rate. 
}
\label{tab:samp_comp}
\begin{threeparttable}
\renewcommand\arraystretch{1}
\setlength\tabcolsep{5pt}
\fontsize{7}{8}\selectfont\begin{tabular}{lcccccccc}
\toprule
& \multicolumn{4}{c}{\textbf{Training Environment}} & \multicolumn{4}{c}{\textbf{Testing Environment}} \\
\cmidrule(lr){2-5} \cmidrule(lr){6-9}
\multirow{-0.30}{*}{\textbf{Method}} 
& \makecell{Soln. Time  \\ {[s]} $\downarrow$}
& \makecell{Path Length  \\ {[m]} $\downarrow$}
& \makecell{Succ. Rate  \\ {[\%]} $\uparrow$}
& \makecell{Coll. Rate  \\ {[\%]} $\downarrow$}
& \makecell{Soln. Time  \\ {[s]} $\downarrow$}
& \makecell{Path Length  \\ {[m]} $\downarrow$}
& \makecell{Succ. Rate \\ {[\%]} $\uparrow$}
& \makecell{Coll. Rate \\ {[\%]} $\downarrow$} \\
\midrule

cuRobo (Opti.) \cite{cuRobo} & 7.9$e-$1 $\pm$ 3.2$e-$1 & 0.7 $\pm$ 0.2 & 100 & 0 & 8.1$e-$1 $\pm$ 3.3$e-$1 & 0.8 $\pm$ 0.3 & 100 & 0 \\
cuRobo (Sampling) \cite{cuRobo} & 6.0$e-$1$\pm$ 0.6$e-$1 & 1.2 $\pm$ 0.4 & 87 & 0 & 6.1$e-$1 $\pm$ 0.5$e-$1 & 1.3 $\pm$ 0.4 & 86 & 0 \\
\midrule
MPNet \cite{MPN} & 5.0$e-$1 $\pm$ 3.0$e-$1 & 3.2 $\pm$ 0.9 & 65 & 25 & 5.1$e-$1 $\pm$ 2.6$e-$1 & 2.9 $\pm$ 0.7 & 55 & 37 \\
SIMPNet \cite{simpnet} & 4.5$e-$1 $\pm$ 2.0$e-$1 & 2.8 $\pm$ 1.7 & 75 & 20 & 4.1$e-$1 $\pm$ 2.3$e-$1 & 2.8 $\pm$ 0.9 & 40 & 47 \\
MPiNet \cite{motionpolicynetwork} & 5.4$e-$1$\pm$2.4$e-$1 & 0.9 $\pm$ 0.3 & 80 & 13 & 5.7$e-$1$\pm$1.4$e-$1 &  1.0 $\pm$ 0.4 & 68 & 16 \\
\midrule
SAMP (100, rollout) & \textbf{7.3$\bm{e-}$2 $\bm{\pm}$ 1.7$\bm{e-}$2} & 0.8 $\pm$ 0.4 & 82  &  10 & \textbf{7.4$\bm{e-}$2 $\bm{\pm}$ 4.3$\bm{e-}$2} & 1.0 $\pm$ 0.4 & 75 & 13 \\
SAMP (160, rollout) & 8.7$e-$2 $\pm$ 1.6$e-$2 & \textbf{0.8 $\bm{\pm}$ 0.3} & \textbf{90}  & \textbf{6} & 8.9$e-$2 $\pm$ 1.9$e-$2 & \textbf{0.9 $\bm{\pm}$ 0.3} & \textbf{81}  & \textbf{7} \\
\bottomrule
\end{tabular}
\begin{tablenotes}[para,flushleft]
\footnotesize
Note: As cuRobo (Opti.) is the expert planner that our policy learned, the best result except cuRobo is reported as the best.
\end{tablenotes}
\end{threeparttable}
\end{table*}

\begin{table*}[tbp]
\centering
\begin{threeparttable}
\caption{Ablation study on the impact of spatial anchor size. Inference time (Inf. Time) refers to the computational cost of a single-step policy prediction, while training time (Train. Time) indicates the computation time required for each training epoch.}
\label{tab:aba_anchor_size}
\renewcommand\arraystretch{1}
\setlength\tabcolsep{3pt}
\fontsize{7}{8}\selectfont
\begin{tabular}{cccccccccc}
\toprule
& & \multicolumn{3}{c}{\textbf{Training Environment}} & \multicolumn{3}{c}{\textbf{Testing Environment}} \\
\cmidrule(lr){3-5} \cmidrule(lr){6-8}
\makecell{Anchor Size}
& \makecell{Train. Time {[s]} $\downarrow$}
& \makecell{Inf.  Time {[s]} $\downarrow$}
& \makecell{Succ.  Rate {[\%]} $\uparrow$}
& \makecell{Reach  Rate {[\%]} $\uparrow$}
& \makecell{Coll.  Rate {[\%]} $\downarrow$}
& \makecell{Succ.  Rate {[\%]} $\uparrow$}
& \makecell{Reach  Rate {[\%]} $\uparrow$}
& \makecell{Coll.  Rate {[\%]} $\downarrow$}\\
\midrule
40 & 150 & 8$e-$4 $\pm$ 6$e-$5 & 75 & 79 & 13 & 65 & 76 & 15  \\
70 & 180 & 1.1$e-$4 $\pm$ 1$e-$4 & 78 & 83 & 11 & 70 & 80 & 15  \\
100 & 270 & 1.5$e-$3 $\pm$ 2$e-$4 & 82 & 88 & 10  & 75 & 83 & 13  \\
130 & 560 & 1.8$e-$3 $\pm$ 4$e-$4 & 86 & 91 & 8  & 77 & 88 & 9  \\
160 & 1050 & 2.0$e-$3 $\pm$ 2$e-$4 & \textbf{90} & \textbf{95} & \textbf{6}  & \textbf{81} & \textbf{89} & \textbf{7}  \\
\bottomrule
\end{tabular}
\end{threeparttable}
\end{table*}

\subsection{Ablation Study} 

In this section, a series of ablation studies is conducted to analyze the effects of key parameters and modules in our method. We systematically investigate the effects of anchor point resolution and critical architectural components, using quantitative comparisons in testing environments. These studies demonstrate the impact of each factor on the performance, generalization, and computational efficiency of motion policies. The following subsections provide detailed analyses and insights into the trade-offs and effectiveness of our framework.

\begin{figure}[tbp]
\centering 
\includegraphics[width=0.485\linewidth]{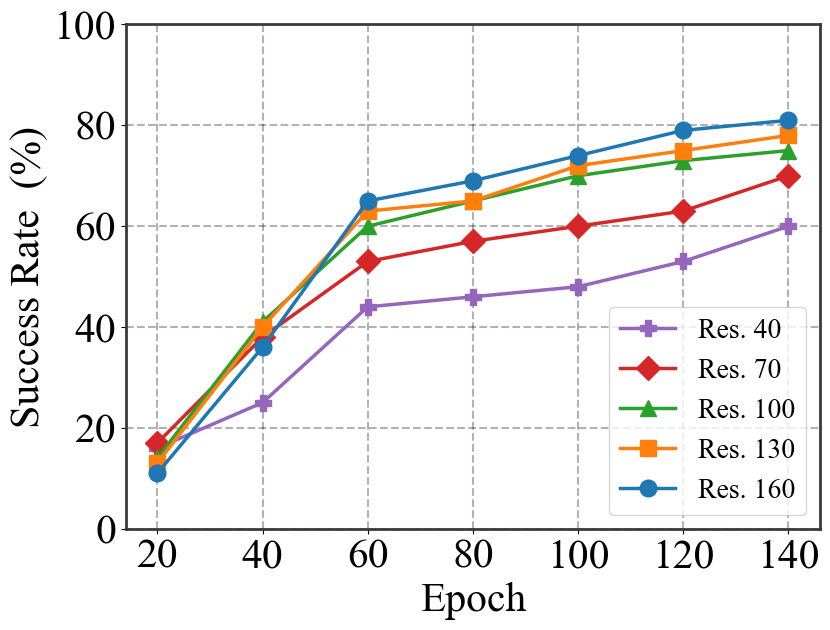}
\includegraphics[width=0.485\linewidth]{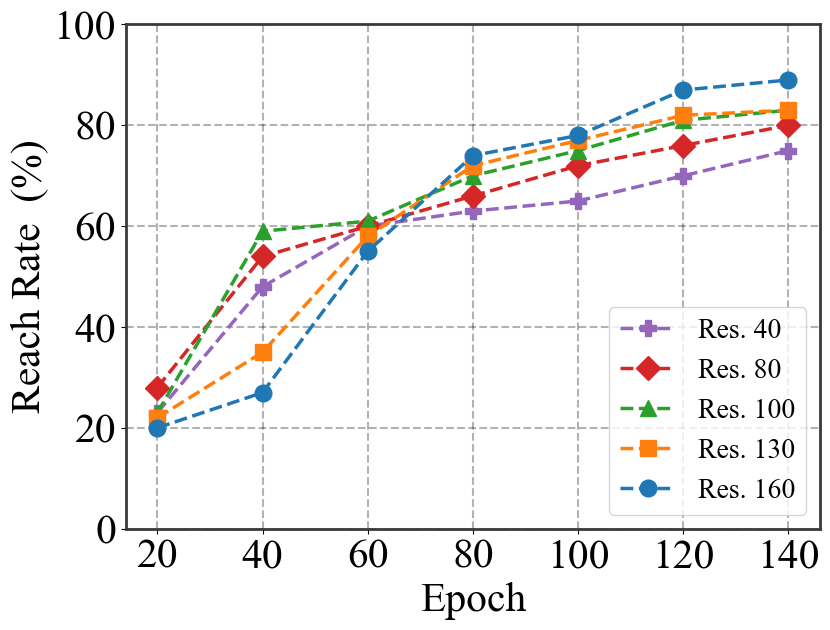}
\caption{Performance comparison of models trained with different anchor point resolutions (Res.), evaluated by success rate (left) and reach rate (right) across training epochs.}
\label{fig:aba_train_epoch}
\end{figure}

\begin{figure}[tbp]
\centering 
\includegraphics[width=0.485\linewidth]{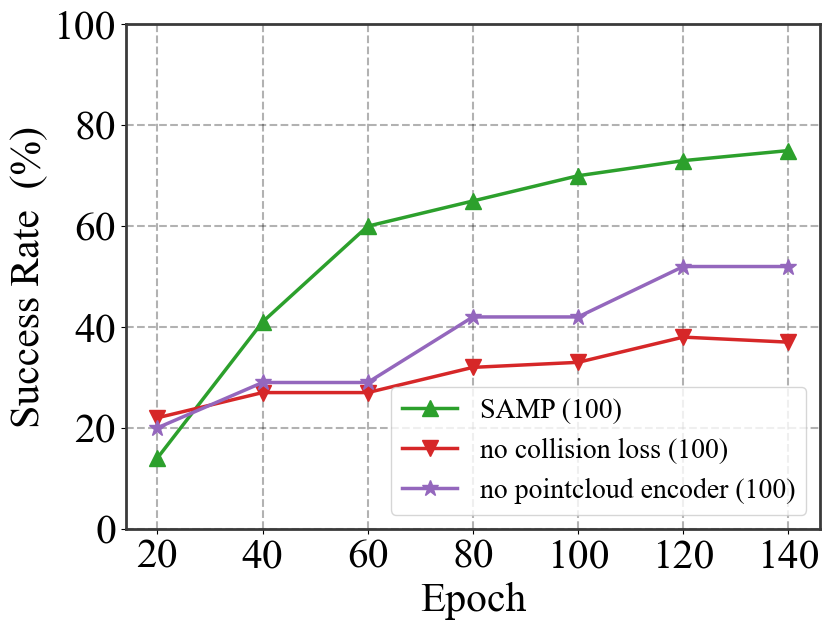}
\includegraphics[width=0.485\linewidth]{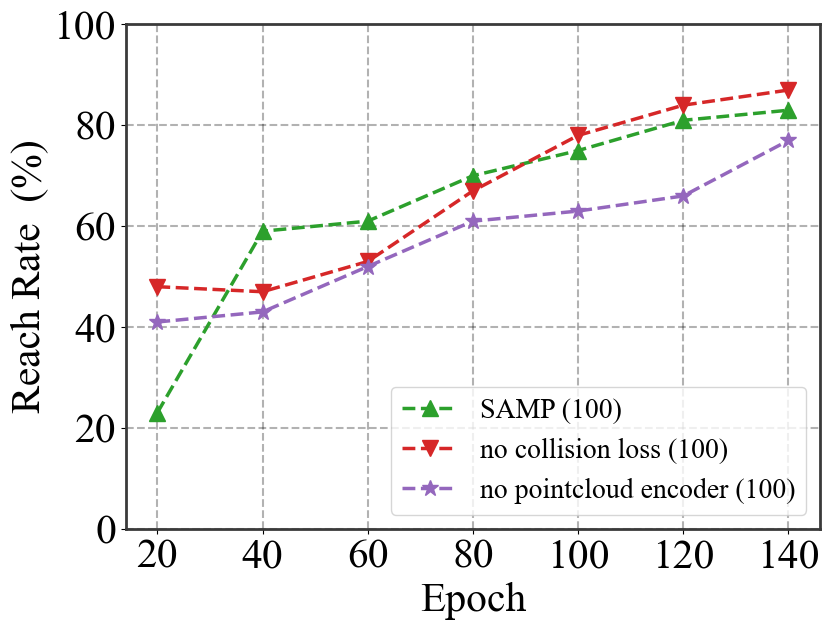}
\caption{Ablation study evaluating the effect of individual components of SAMP on task performance using a resolution of 100 anchor points.}
\label{fig:aba_module}
\end{figure}

\subsubsection{Different Resolutions of the Anchor Points}

The ablation results in Table \ref{tab:aba_anchor_size} demonstrate significant trends in motion planning performance relative to anchor size variations. Increasing the anchor size from $40$ to $160$ units yields substantial improvements across all metrics. Success rates rise from $75$\,\% to $90$\,\% in training environments and from $65$\,\% to $81$\,\% in testing environments. Similarly, reach rates improve from $79$\,\% to $95$\,\% during training and from $76$\,\% to $89$\,\% during testing. Concurrently, collision rates decline from $13$\,\% to $6$\,\% in training and from $15$\,\% to $7$\,\% in testing environments. The narrowing performance gap between training and testing environments particularly in collision reduction suggests that larger anchor sizes enhance model generalization. 

To examine how network performance evolves over training epochs under different anchor point resolutions, we plot the learning curves shown in Fig.~\ref{fig:aba_train_epoch}. As shown in the left plot, the success rate steadily increases with the number of epochs, indicating that the network progressively acquires more reliable planning capabilities. In the early training stage, all models exhibit relatively low success rates, but performance improves rapidly as training proceeds. Beyond $80$ epochs, most models start to plateau, with larger models achieving higher final success rates compared to smaller ones. A similar trend can be observed in the right plot for the reach rate a sharp improvement occurs in the early epochs, followed by gradual refinement as training continues. Notably, deeper models consistently outperform shallower ones in the later training stage, demonstrating that increased capacity helps the network generalize better and converge to higher performance.

These collision avoidance performance gains, however, accompany increased computational demands primarily due to the larger spatial representation processing. The training time rises from $150$\,s to $1050$\,s and inference time growth from $0.8$\,ms to $2.0$\,ms directly results from the expanded anchor size, which increases the computational complexity of both the 3D convolutional operations and the joint grid sampling process. The 3D convolution requires more extensive kernel applications across the dense spatial configurations, while the joint grid sampling must handle a greater number of anchor points for feature alignments.

\begin{figure*}[t]
    \centering
    \includegraphics[width=1.0\linewidth]{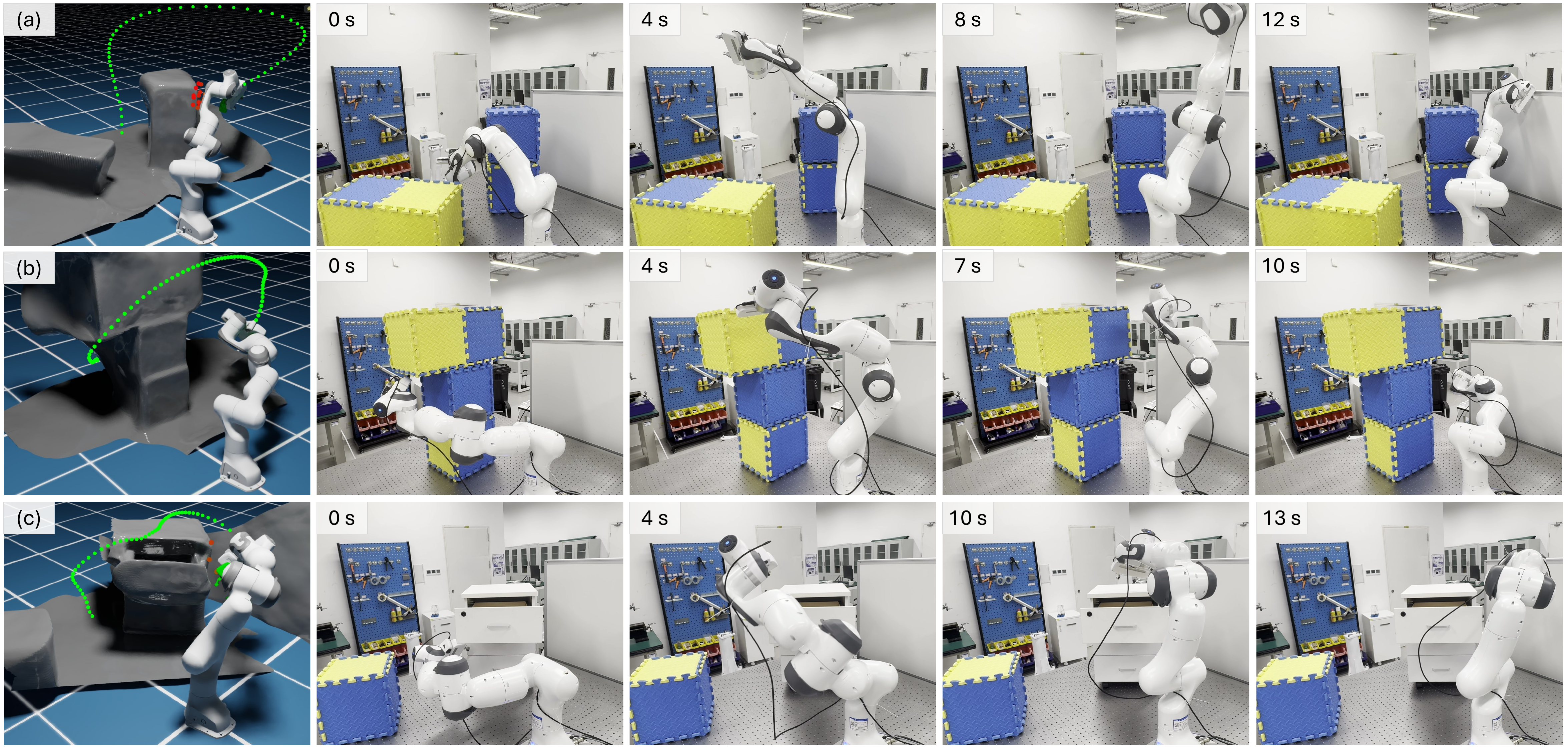}
    \caption{Qualitative results of collision-free motion planning using SAMP in two real scenarios. Each row corresponds to a different environment: (a) Multiple blocks, (b) a single large ``T'' block, and (c) a cabinet with an opening drawer. The leftmost images visualize the planned trajectories, where obstacle geometry is reconstructed from neural-based SDF. The subsequent frames show the real-world execution of the generated trajectory at different time stamps.}
    \label{fig:exp_real}
\end{figure*}

\subsubsection{Network Architecture and Loss Function}

To assess the contribution of key components in our framework, we perform ablation studies by selectively removing the collision loss and the point cloud encoder. As illustrated in Fig.~\ref{fig:aba_module}, omitting the collision loss results in a noticeable decrease in success rate throughout training, highlighting the importance of explicit collision modeling for learning collision-free behaviors. Likewise, removing the point cloud encoder significantly degrades both the success rate and reach rate, underscoring the critical role of geometric context in guiding the policy toward feasible motions. In contrast, our complete method consistently achieves the highest success and reach rates as training progresses, although the incomplete variants may initially converge faster due to the simplified setting. These experiments collectively demonstrate the synergistic effects of these components in enhancing overall performance.

\subsection{Real-World Experiments} 

To further evaluate the applicability of SAMP in real-world settings, we conduct a series of experiments with a physical robot manipulator in various cluttered environments. For each scenario, we first capture depth images of the scene using an RGB-D sensor, and reconstruct the obstacle geometry to obtain a neural SDF representation of the workspace. This learned SDF model is then provided for SAMP to infer the distance values of the anchor points as the environment representation. During the experiment, the robot iteratively generates and follows collision-free trajectories based on this SDF, until the target pose is reached. The resulting motion plans are executed directly on the real robot without the need for additional post-processing or manual intervention. As shown in Fig. \ref{fig:exp_real}, SAMP enables the manipulator to smoothly and safely reach its target poses within complex and constrained environments. These results demonstrate that SAMP can robustly operate on neural SDFs reconstructed from real sensor data, enabling reliable and efficient motion planning in complex, unstructured settings.





\section{Conclusion and Future Work}

In this paper, we introduce SAMP, a novel motion policy framework that efficiently plans collision-free robot trajectories by integrating spatial anchor points with SDF representations in the workspace. The method leverages neural implicit SDF to achieve unified and accurate encoding of both environmental structures and robot geometry, enabling precise collision inference without expensive explicit checks. 
Through extensive experiments, we demonstrate that SAMP exhibits strong adaptability and robust planning performance across a range of diverse and complex scenarios. Notably, SAMP generalizes effectively to unseen target positions and novel environments, outperforming existing methods that rely on hand-crafted geometric priors or discretized representations. These results highlight the potential of combining neural representations with principled motion planning frameworks for efficient robot motion planning. In addition, our real-world experiments further validate SAMP’s effectiveness and demonstrate its promise for deployment in actual robotic systems, paving the way for reliable application in real environments. In future work, we will explore advanced network architectures to enhance SAMP’s capacity for feature extraction and trajectory generation in challenging settings. Furthermore, we aim to employ reinforcement learning techniques to fine-tune the motion policy, which is expected to improve adaptability and generalization through experience-driven optimization.

\bibliographystyle{IEEEtran}
\bibliography{mybibs}

\end{document}